# Grid-based angle-constrained path planning

Konstantin Yakovlev, Egor Baskin, Ivan Hramoin

Institute for Systems Analysis of Russian Academy of Sciences, Moscow, Russia
{yakovlev, baskin, hramoin}@isa.ru

**Abstract.** Square grids are commonly used in robotics and game development as spatial models and well known in AI community heuristic search algorithms (such as A*, JPS, Theta* etc.) are widely used for path planning on grids. A lot of research is concentrated on finding the shortest (in geometrical sense) paths while in many applications finding smooth paths (rather than the shortest ones but containing sharp turns) is preferable. In this paper we study the problem of generating smooth paths and concentrate on angle constrained path planning. We put angle-constrained path planning problem formally and present a new algorithm tailored to solve it – LIAN. We examine LIAN both theoretically and empirically. We show that it is sound and complete (under some restrictions). We also show that LIAN outperforms the analogues when solving numerous path planning tasks within urban outdoor navigation scenarios.

## 1 Introduction

Path planning is one of the key abilities for an intelligent agent (robot, unmanned vehicle, computer game character etc.) to autonomously operate in either real or virtual world. Typically, in Artificial Intelligence, agent's environment is modeled with weighted graph which vertices correspond to locations the agent can occupy and edges correspond to trajectories the agent can traverse (line segments, curves of predefined shape etc.). Each edge is assigned a non-negative real number (weight, cost) by a weighting function which is used to quantitatively express characteristics of the corresponding trajectory (length, potential risk of traversing, etc.). Thus to solve a path planning problem one needs *a)* to construct a graph (given the description of the environment) and *b)* to find a path (preferably – the shortest one) on this graph.

Among the most commonly used graph models one can name visibility graphs [1], Voronoi diagrams [2], navigation meshes [3], regular grids [4]. The latter are the most widespread for several reasons. First, they appear naturally in many virtual environments (computer games are the most obvious example [5]) and even in real world scenarios, say in robotics, it is the grids that are commonly used as spatial models [6]. Second, even if the environment is described in some other way it is likely that forming a grid out of this description will be less burdensome than constructing other abovementioned models due to grid's "primitive" structure.

After the graph is constructed the search for a path on it can be carried out by the well known Dijkstra's algorithm [7] or A* algorithm [8] (which is the heuristic modification of Dijkstra) or many of their derivatives: ARA* [9], HPA* [10], R* [11], Theta* [12], JPS [13] to name a few. Some of these algorithms are tailored to grid path finding (JPS, Theta*, HPA*), others are suitable for any graph models (with A* and Dijkstra being the most universal ones). Many of them, in fact – almost all of them, overcome their predecessors in terms of computational efficiency (at least for a large class of tasks). Some algorithms are tailored to single-shot path planning while others demonstrate their supremacy on solving bunches of tasks. But only a few of them are taking the shape of the resultant path into account although it can be quite useful in many applications. For example, a wheeled robot or an unmanned aerial vehicle simply can not follow a path with sharp turns due to their dynamic constraints. The most common way to incorporate these constraints into path planning process is to extend the search space with the agent's control laws encodings – see [14] for example. This leads to significant growth of the search space and path finding becomes computationally burdensome. So it can be beneficial to stay within grid-based world model and spatial-only search space and focus on finding the smooth paths (rather than the short ones) and thus indirectly guarantee the feasibility of that paths against the agent's dynamic constraints.

We find the idea of generating smooth paths very appealing and address the following angle constrained path planning problem. Given a square grid the task is to find a path as a sequence of grid sections (ordered pairs of grid elements) such that an angle of alteration between each two consecutive sections is less or equal than some predefined threshold. We present novel heuristic search algorithm – LIAN (from "limited angle") – of solving it. We examine LIAN both theoretically, showing that it is sound and complete (under some constraints), and experimentally, testing LIAN's applicability in urban outdoor navigation scenarios.

To the best of our knowledge, no direct competitors to LIAN are present nowadays, although there exists one or more implicit analogues – path planning methods that can be attributed to as taking the shape of the path into account. For example A*-PS [10] runs A*-search on a grid and after it is finished performs a preprocessing step to eliminate intermediate path elements. Thus the resultant path starts looking more realistic and at the same time it becomes shorter. Theta* (or more precise – Basic Theta*) [12] uses the same idea – intermediate grid elements skipping – but it performs the smoothing procedure online, e.g. on each step of the algorithm. In [15] a modification of Basic Theta* (also applicable to A*-PS) algorithm is presented which uses special angle-based heuristic to focus the search in order to construct more straightforward paths to the goal. In [16] another modification of Basic Theta* – weighted angular rate constrained Theta* (wARC-Theta*) - is described. wARC-Theta* uses special techniques to take into account agent's angular rate (and other) constraints staying within grid model e.g. without extending the spatial model with agent's orientation (heading) information but rather performing the corresponding calculations online. wARC-Theta* with some minor adaptations can be used to solve the angle constrained path planning problem we are interested in. Unfortunately, the algorithm is incomplete, e.g. it fails to solve a wide range of path planning tasks alt-hough the solutions to these tasks do exist. With some modifications, explained further in the paper, the performance of wARC-Theta* can be improved and the number of successfully solved tasks can be increased. This improved version of wARC-Theta* is seen to be the only direct analogue of the proposed algorithm so we use it to perform the comparative experimental analysis. Obtained results show that the newly proposed algorithm – LIAN – significantly outperforms wARC-Theta*: LIAN solves much more tasks and uses significantly less computational resources (processor time and memory).

The latter of the paper is organized as follows. In section 2 we express the angle constrained path planning problem formally. In section 3 the new algorithm of solving it – LIAN – is present, as well as modified wARC-Theta* algorithm is described. In section 4 the results of the comparative experimental study are given.

## 2    Angle constrained path planning problem on square grid

Two alternative types of square grid notations are widespread nowadays: center-based, when agent's locations are tied to the centers of grid cells, and corner-based, when agent's locations are tied to the corners, respectively (see figure 1).

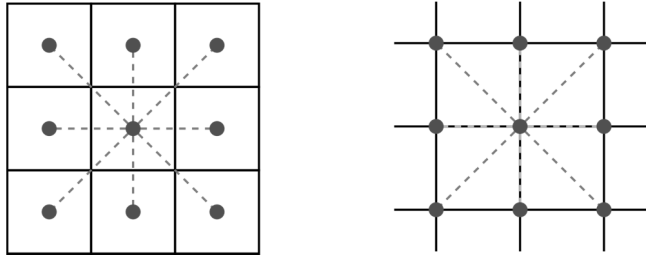

**Fig. 1.** Square grids: center-based (left) and corner-based (right).

We adopt the center-based notation and consider a grid to be a finite set of cells $A$ that can be represented as a matrix $A_{M \times N} = \{a_{ij}\}$, where $i, j$ – are cell position indexes (coordinates) and $M, N$ – are grid dimensions. Each cell is labeled either traversable or un-traversable and the set of all traversable cells is denoted as $A^+$. In case cell coordinates can be omitted, lower case Latin characters will be used: $a, b, c$ etc.

A line-of-sight function, $los$: $A^+ \times A^+ \rightarrow \{true, false\}$, is given and an agent is allowed to move from one traversable cell to the other if $los$ returns true on them (or, saying in other words, if there exist a line-of-sight between them). In our work, as in many others, we use well-known in computer graphics Bresenham algorithm [17] to detect if line-of-sight between two cells exist or not. This algorithms draws a "discrete line section" (see figure 2) and if it contains only traversable cells than $los$ is supposed to return true (otherwise $los$ returns false).

A metric function, $dist$: $A^+ \times A^+ \rightarrow \Re$, is given to measure the distance between any two traversable cells. We use Euclid distance, e.g $dist(a_{ij}, a_{lk}) = \sqrt{(l-i)^2 + (k-j)^2}$ as metric function.

An ordered pair of grid cells is a section: $e=\langle a_{ij}, a_{lk}\rangle$, and it is traversable *iff los*($a_{ij}$, $a_{lk}$)=*true*. The length of a section $\langle a_{ij}, a_{lk}\rangle$ equals *dist*($a_{ij}, a_{lk}$). Two sections that have exactly a middle cell in common, e.g. $e_1=\langle a_{ij}, a_{lk}\rangle$, $e_2=\langle a_{lk}, a_{vw}\rangle$, are called adjacent.

$\Delta$-section is such a section $e=\langle a_{ij}, a_{lk}\rangle$ that it's endpoint, $a_{lk}$, belongs to *CIRCLE*($a_{ij}$, $\Delta$), where *CIRCLE* is a set of cells identified by the well-known in computer graphics Midpoint algorithm [18] (which is a modification of the abovementioned Brezenham's algorithm for drawing "discrete circumferences") – see figure 2.

A path between two distinct traversable cells *s* (start cell) and *g* (goal cell) is a sequence of traversable adjacent sections such that the first section starts with *s* and the last ends with *g*: $\pi(s, g)=\pi=\{e_1, ..., e_v\}$, $e_1=\langle s, a\rangle$, $e_v=\langle b, g\rangle$. The length of the path *len*($\pi$) is the sum of the lengths of the sections forming that path.

Given two adjacent sections $e_1=\langle a_{ij}, a_{lk}\rangle$, $e_2=\langle a_{lk}, a_{vw}\rangle$ an angle of alteration is the angle between the vectors $\overrightarrow{a_{ij}a_{lk}}$ and $\overrightarrow{a_{lk}a_{vw}}$, which coordinates are ($l - i, k - j$) and ($v - l, w - k$) respectively (see figure 2). This angle is denoted as $\alpha(e_1, e_2)$ and it's value is denoted as $|\alpha(e_1, e_2)|$.

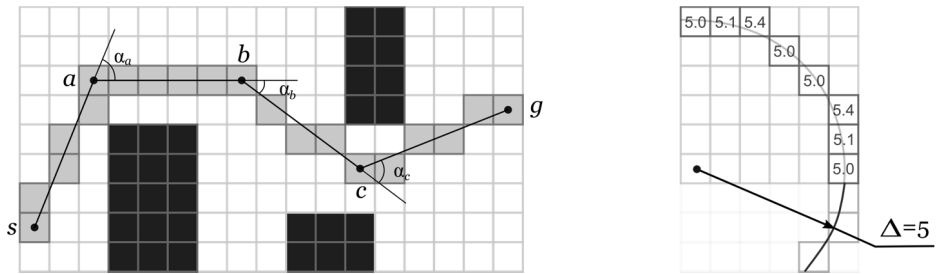

**Fig. 2.** Main concepts of the angle constrained path planning problem. On the left: traversable sections are depicted as solid lines; cells identified by the Bresenham algorithm are shaded grey; angles of alterations are denoted as $\alpha_a, \alpha_b, \alpha_c$; the path being depicted is a $\Delta$-path, $\Delta$=5. On the right: *CIRCLE* set – cells identified by the Midpoint algorithm, $\Delta$=5.

Given a path $\pi=\{e_1, ..., e_v\}$ we will call the value $\alpha_m(\pi)=\alpha_m=\max\{|\alpha(e_1, e_2)|, |\alpha(e_2, e_3)|, ..., |\alpha(e_{v-1}, e_v)|\}$ the maximum angle of alteration of the path.

Now we are interested in solving angle constrained path planning problem which is formulated as following. Given two distinct traversable cells *s* (start cell) and *g* (goal cell) and the value $\boldsymbol{\alpha_m}$: $0<\boldsymbol{\alpha_m}<180$, find a path $\pi(s, g)$ such that $\alpha_m(\pi)\leq\boldsymbol{\alpha_m}$ (angle constrained path).

Shortest angle constrained path is considered to be the optimal solution. For the reasons explained further in the paper, we are also interested in a special class of solutions of the problem, so called $\Delta$-solutions. $\Delta$-solution is an angle constrained path each section of which, except maybe the last one, is the $\Delta$-section (the path depicted on the figure 2 is a $\Delta$-path, $\Delta$=5).

## 3 Algorithms for the angle constrained path planning

### 3.1 wTheta*-LA

In [13] H. Kim *et al.* present a modification of Basic Theta* [9] algorithm tailored to solve grid path planning problem for an agent with angular rate constraints. Authors do not consider the maximum angle of alteration constraint – as described above – directly. Instead, they investigate the case when the speed and the turning radius of an agent are given and calculate angle constraint online, taking into account the length of the path sections involved. But if one replaces the original procedure of angle constraint calculation with the one which always returns $α_m$, their algorithm becomes applicable to the angle constrained path problem we are interested in. We call such an algorithm Theta*-LA (LA stands for "limited angle").

Theta*-LA is a pretty straightforward modification of Theta*. The only difference is that when Theta* tries to connect a cell to it's grandparent (in order to skip the intermediate element, e.g. parent, from the path) it validates only the line-of-sight constraint (e.g. if line-of-sight exists between the cell and it's grandparent the former is being connected to the latter), while Theta*-LA validates also angle constraint, and if an angle between the sections defined by the trio: grandparent-parent-cell is greater than the predefined threshold $α_m$, than parent cell is kept in the sequence. This straightforward technique leads to the following problem: if the angle constraint is less than 45° (which is likely to be a common, realistic scenario) the algorithm fails to circumnavigate large obstacles and thus fails to find an angle constrained path - see figure 3 for detailed explanation.

The main reason Theta*-LA fails to find a path in many cases is that it does not store the intermediate path elements but rather tries to make path sections as long as possible. In the original paper [13] H.Kim *et. al* give a hint how this problem can be partially solved but do not describe it in details – they suggest weighting the grid, e.g. assigning each grid cell a non-negative weight value and taking cells' weights into account while calculating the length of the section. Using weights to penalize the cells residing close to the obstacles in such way that Theta*-LA first prefers processing cells residing at some distance of the obstacles potentially leads to another grandparent-parent-cell sequences and improves the overall performance of the algorithm.

We have implemented the grid weighting procedure that makes cells lying close to the obstacles less attractive to the algorithm and call such an algorithm wTheta*-LA. We use the following strategy: given two parameters – radius *r* and max weighting penalty *p* – discrete circumferences of radius *r* with the centers in the cells *a* lying on the boundaries of the obstacles are constructed (by the referred in section 1 Midpoint algorithm). Than the rays connecting *a* and each cell forming the circumference are traced and each ray cell, say *a'*, is assigned the weight as follows: $w(a') = p \cdot (1 + (1 - dist(a, a'))/r)$ - see figure 3. During the search, a modified length calculation formula is used, e.g. $len(\langle a, b \rangle) = dist(a, b) \cdot (1 + avgW)$, where *avgW* – is the average weight of the cells lying on Bresenham line in between *a* and *b*.

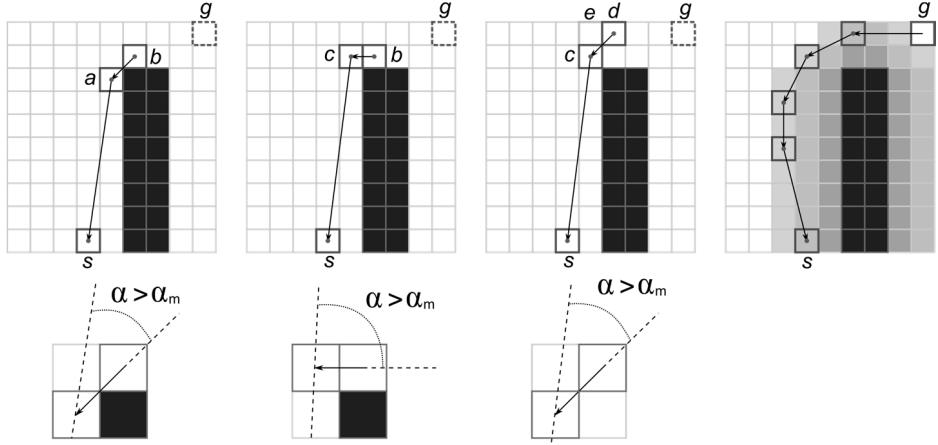

**Fig. 3.** Theta*-LA and wTheta*-LA circumnavigating the obstacle. Left: when expanding cell *a*, *b* is discarded due to the violation of the maximum angle alteration constraint, the search continues to *c*; *b* is discarded again for the same reason; *d* is also discarded so the search would continue to *e*, which is a dead end. Right: weighting the grid alters the direction of the search and the obstacle can be successfully circumnavigated.

Experimental analysis (see section 4) shows that weighting significantly improves algorithm's performance, but still vast variety of path planning tasks remains unsolved. One can suggest playing further with the weighting parameters values or modifying the weighting procedure itself, but we prefer to design a new algorithm that a) does not require any grid preprocessing at all and b) theoretically guarantees completeness (at least for a defined class of tasks). Such an algorithm is described further.

### 3.2 LIAN

LIAN (from "limited angle") is a heuristic search algorithm tailored to solve angle constrained path planning problem on square grids. LIAN relies on A* [5] search strategy of the state-space, uses line-of-sight checks as Theta* [9] and exploits the idea of multiple parents as R* [8].

As well as A* our algorithm explores the grid cells and calculates so called *g*-values, where *g*-value of a cell *a*, $g(a)$, is the length of the path (angle constrained path) from start cell *s* to *a* found so far. Along with the *g*-value each cell is obligatory characterized by the parent pointer (like Theta* but unlike A* where parent pointers are commonly used but are not obligatory) – $bp(a)$, which points to the grid cell which is a predecessor of *a*. Any grid cell can have *multiple* parents (this makes LIAN similar to R*). Thus when we are talking about the search space of LIAN we are talking about the space consisting of nodes which are the triples: cell, *g*-value, parent pointer (which actually points to the node, not the cell). Nodes will be denoted as [**a**], and [**a**]=[*a*, *g*([**a**]), *bp*([**a**])]. So, [**a**] is a node and *a* is a corresponding cell. *bp*([**a**]) is a node (e.g. *bp*([**a**])=[**a'**]) and *bp*(*a*) is a corresponding cell (*bp*(*a*)=*a'*).

As well as any other A*-like search algorithm LIAN maintains and stores in memory two lists of nodes: *OPEN* and *CLOSED*. *OPEN* is the set of nodes – potential candidates for further processing and it initially contains the only element [*s*, 0, ∅]. *CLOSED* is the set of nodes that have already been processed. On each step node [**a**] with minimal *f*-value, *f*([**a**]),is retrieved from *OPEN*, where *f*([**a**])=*g*([**a**])+*h*(*a*), and *h*(*a*) is a heuristic estimate (e.g. *dist*(*a*, *g*)) of the path length from *a* to the goal cell (as in A*). Then the potential successors of [**a**] are generated *SUCC*([**a**])=*SUCC*. In A* *SUCC* is formed out of the cells which are adjacent to *a*. Unlikely, in LIAN potential successors correspond to the cells residing at the fixed distance Δ∈*N* (which is the input parameter of the algorithm) from *a*. To identify such cells Midpoint algorithm (described in section 1) is invoked: a discrete circumference of radius Δ is drawn and the nodes corresponding to the cells forming this circumference are added to *SUCC*. If the distance from *a* to *g* is less than Δ, then the goal node is also added to the *SUCC* list. To distinguish between the potential successor nodes and the corresponding cells we will use the record [**succ**$_i$] to denote the former and *succ*$_i$ to denote the latter.

After the set of potential successors is constructed it's pruning is done. First nodes corresponding to un-traversable cells are eliminated. Second, the nodes that violate line-of-sight constraint are pruned. Third, the nodes that correspond to the cells that violate maximum angle of alteration constraints are discarded, e.g. the nodes [**succ**$_i$] that correspond to such cells *succ*$_i$: |α(⟨*bp*(*a*), *a*⟩, ⟨*a*, *succ*$_i$⟩)|>*α*$_m$ (*NB*: if the start node is processed the angle constraints are ignored). Forth, the cells that have been visited before are pruned, e.g. if the *CLOSED* list contains a node with the same cell and parent pointer then such potential successor is discarded.

```
1.   LIAN(start, goal, Δ, αₘ)                      14.  Expand([a], Δ, αₘ)
2.     bp([start]) := ∅; g([start]) := 0;          15.    SUCC = getCircleSuccessors([a], Δ);
3.     OPEN.push([start]); CLOSED := ∅;            16.    if dist(a, goal) < Δ
4.     while OPEN.size > 0                         17.      SUCC.push([goal]);
5.       [a] := argmin[a]∈OPEN f([a]);             18.    for each [a'] ∈ SUCC
6.       OPEN.remove([a]);                         19.      if a' is un-traversable
7.       if a = goal                               20.        continue;
8.         getPathFromParentPointers([a]);         21.      if | α(⟨bp(a), a⟩, ⟨a, a'⟩) | > αₘ
9.         return "path found";                    22.        continue;
10.      CLOSED.push([a]);                         23.      for each [a"] ∈ CLOSED
11.      Expand([a], Δ, αₘ);                       24.        if a'=a" and bp(a')=bp(a")
12.    return "no path found"                      25.          continue;
13.  end                                           26.      if LineOfSight(a, a') = false
                                                   27.        continue;
                                                   28.      g([a']) := g([a]) + dist(a, a');
                                                   29.      OPEN.push([a']);
                                                   30.  end
```

**Fig. 4.** LIAN Algorithm

After fixing the *SUCC* set, *g*-values of the successors are calculated: $g([succ_i])=g([a])+d(a, succ_i)$ and corresponding nodes are added to *OPEN*. [**a**] is added to *CLOSED*.

Algorithm's stop criterion is the same as used in A*: LIAN stops when a node corresponding to the goal cell is retrieved from *OPEN* (in that case the path can be reconstructed using parent pointers). If the *OPEN* list becomes empty during the search algorithm reports *failure* to found a path.

The proposed algorithm has the following properties.

**Property 1**. LIAN always terminates.
**Sketch of proof**. Algorithm is performing the search until the *OPEN* list is empty (or until the goal node is retrieved from it). *OPEN* contains only elements that refer to the grid cells the total number of which is finite. The number of potential parents of the cell is also finite. At the same time when a new node is generated LIAN checks whether this node (the node defined by the same cell and the same parent) has been processed before already (lines 24-26). And in case the answer is 'yes' it is pruned and not added to *OPEN*. Thus, the total number of nodes potentially addable to *OPEN* is finite. Given the fact that on each step of the algorithm an element is removed from *OPEN* (line 6) one can infer that sooner or later this list will contain no elements, or the goal node will be retrieved. In both cases (lines 4, 7) algorithm terminates.

**Property 2**. If only Δ-solutions are under investigation then LIAN is sound and complete, e.g. if Δ-solution to the angle constrained path planning task exists LIAN finds it, if no Δ-solution exist, LIAN reports *failure*.
**Sketch of proof**. LIAN's parameter Δ well defines the set of potential successors for any node as the set of nodes corresponding to the cells residing at the Δ-distance. All successors that correspond to the traversable cells and satisfy the maximum angle of alteration and line-of-sight constraints are added to *OPEN* (except those that have been examined before). Thus, sooner or later *all* paths compromised of the Δ-sections (except, maybe, the last section – lines 16-17) will be constructed and evaluated and the sought path, if it exist, will be found. By construction this path is a Δ-solution of the given task. If LIAN reports *failure* it means that *all* the potential paths – candidates for the Δ-solution have been examined (otherwise *OPEN* list still contains some elements and LIAN continues the search), which in turn means no Δ-solution exists.

**Property 3**. If different Δ-solutions to the angle constrained path planning task exist LIAN returns the shortest one.
**Sketch of proof**. LIAN uses the same *OPEN* prioritization strategy as A* which guarantees finding the shortest path if the admissible heuristic is used. LIAN uses Euclidian distance function *dist*, which is admissible (and consistent as well) heuristic. Thus LIAN returns the shortest Δ-solution possible.

We would like to notice further that just like A* LIAN allows heuristic weighting, e.g. calculating *f*-values in the following way – $f([a])=g([a])+w \cdot h(a)$, where $w>1$. Weighting the heuristic commonly makes it inadmissible thus the optimality of the solution can not be guaranteed any more. But at the same time, it's known that in many practical applications, grid path planning inclusively, heuristic weighting radi-cally improves algorithm's performance while the quality of the solution decreases insignificantly. So we would also like to use LIAN with weighted heuristic as practically-wise we are interested in finding the solution as quickly as possible.

### 3.3 D-LIAN

Necessity to initialize LIAN with fixed $\Delta$ leads to the obvious problem: which exact value to choose? In cluttered spaces setting $\Delta$ too high will likely make LIAN report *failure* because line-of-sight constraints will be continuingly violated resulting in exhausting of *OPEN* list (there simply will be no candidates to fill it up). At the same time setting $\Delta$ too low leads to the reduction of potential successors set – *SUCC* – for any node under expansion (the lower the value $\Delta$ is the fewer cells form the discrete circumference of radius $\Delta$) and thus *OPEN* list is likely to exhaust again.

To address this problem and make LIAN behavior more flexible and adaptable we suggest dynamically change $\Delta$ while performing the search. The modification of LIAN that uses this technique will be referred to as D-LIAN.

D-LIAN works exactly the same as LIAN but uses a bit modified *Expand*() procedure: it refines the *SUCC* set in 2 phases. Traversability check, maximum angle of alteration check and *CLOSED* list check (lines 19-26) are separated from the line-of-sight check (line 27). Namely, when some [**succ**$_i$] passes checks encoded in lines 19-26 it is added to *SUCC*2 and iteration over *SUCC* set continues. Thus phase 1 ends with forming *SUCC*2 – set of traversable nodes not processed before and not violating maximum angle of alteration constraint. Then all elements of *SUCC*2 are checked against line-of-sight constraint and elements that successfully pass this check are added to *OPEN* (just as before). The difference is when *all* the line-of-sights checks on *SUCC*2 elements fail. In that case $\Delta$ value is half decreased and *Expand*() procedure is invoked again (while usual LIAN just finishes node's expansion and no successors are added to *OPEN*). D-LIAN consequently repeats the *Expand*() procedure (and each time half-decreased value of $\Delta$ is used) until some valid successor(s) is generated or until value of $\Delta$ reaches some predefined threshold – $\Delta_{min}$ (set by the user). In the latter case D-LIAN stops node expansion and no successors are added to *OPEN*.

If, at some step of node [**a**] expansion process, valid successors are generated, $\Delta$-value is remembered and then the search from [**a**] continues using that exact value of $\Delta$ (we will refer to it as to $\Delta([\mathbf{a}])$). If next $n$ successive expansions of [**a**] are all characterized by successful successors generation and decreasing of $\Delta([\mathbf{a}])$ was not used to generate them then $\Delta([\mathbf{a}])$ is half increased. The upper limit on $\Delta$ value – $\Delta_{max}$ – is also set by the user. Thus while performing the search D-LIAN dynamically adjusts $\Delta$ in order to generate as many successors of each node as it is needed to solve the task.

One of the features of D-LIAN is that multiple $\Delta$ values are potentially used during the search. Technically this is achieved by storage of $\Delta$-value referenced to a node. Thus D-LIAN node becomes a quadruple: [$a$, $g(a)$, $bp([\mathbf{a}])$, $\Delta([\mathbf{a}])$]. Input parameters of D-LIAN are: $\Delta_{init}$ – initial value of $\Delta$, $\Delta_{min}$ – the lower threshold, $\Delta_{max}$ – the upper threshold, $n$ – the number of steps after which $\Delta$ is half-increased. In the experiments we used the following bindings: $n=2$, $\Delta_{min}=\Delta_{init}/2$, $\Delta_{max}=\Delta_{init}$.

## 4  Experimental analysis

The experimental setup for the comparative study of the algorithms considered in the paper – LIAN, D-LIAN, Theta*-LA, wTheta*-LA – was the Windows7-operated PC, iCore2 quad 2.5GHz, 2Gb RAM. All the algorithms were coded in C++ using the same data structures and programming techniques.

Urban outdoor navigation scenario was targeted and path finding for small unmanned aerial vehicle (UAV) performing nap-of-the-earth flight was addressed.

Each grid involved in the tests was constructed using OpenStreetMaps (OSM) data [19]. To generate a grid a 1347m x 1347m fragment of actual city environment was retrieved from OSM and discretized to 501 x 501 grid so one cell refers to (approx.) 2,7m x 2,7m area. Cells corresponding to the areas occupied by buildings were marked un-traversable. 80 different city environments were used and 5 different start-goal locations were chosen for each environment fragment residing more than 1350m one from the other (so $dist(start, goal) \geq 500$). Thus, in total, testbed consisted of the 400 various path planning tasks. Targeted angle constraints were: 20°, 25° and 30° (these figures were advised by the peers involved in UAV controllers design).

The following indicators were used to compare the algorithms:

$sr$ – success rate – number of the successfully accomplished angle constrained path planning tasks divided by the number of all tasks;

$t$ – time (in seconds) – time needed for an algorithm to produce solution;

$m$ – memory (in nodes) – number of elements stored in OPEN$\cup$CLOSED (the memory consumption of the algorithm);

$pl$ – path length (in meters) – the length of the resulting angle-constrained path.

Preliminary tests had been conducted to roughly assess the performance of the algorithms. The following observations were made. First, LIAN under some parameterizations terminates minutes after it was invoked. So a 60-seconds time limit was suggested for further testing, e.g. if any algorithm did not terminate within 60 seconds the result of the test was considered to be *failure*. Second, using weighted heuristic radically improves LIAN's computational performance while path length reduces insignificantly (around 1-2%). So if further tests LIAN was run with the heuristic weight equal to 2. Third, "the best" parameters for wTheta*-LA ($p$=0.1, $r$=12) were identified and these parameters were used further on.

The main series of tests involved the following algorithms: 4 instances of LIAN, each using it's own Δ: 3, 5, 10, 20, referred, further as LIAN-3, LIAN-5, LIAN-10, LIAN-20; Theta*-LA and wTheta*-LA. Thus, 7*400=2800 experiments in total were conducted. Obtained results are shown on figure 5.

Figures shown in the table (except $sr$ and *PAR-10* indicators) are the averaged values with *failures* not considered while averaging. Namely, for each algorithm $t$, $m$, $pl$ values were averaged taking into account only it's respective positive results. *PAR-10* is the penalized average runtime – metrics that averages the runtime but takes *failures* into account [20]: if an algorithm fails to solve a task, $t$ is set (penalized) for that run to be 10\**cut-of-time* (where *cut-off-time* equals 60) and in the end all the obtained $t$ values are averaged. Thus *PAR-10* can be seen as an integral indicator of algorithm's ability to solve various path planning tasks as quick as possible.

|        | $\alpha_m = 20$ |||||  $\alpha_m = 25$ ||||| $\alpha_m = 30$ |||||
|--------|----|-----|----|------|------|----|-----|----|------|------|----|-----|----|------|------|
|        | sr | PAR-10 | t | m* | pl | sr | PAR-10 | t | m* | pl | sr | PAR-10 | t | m* | pl |
| **LIAN-3**  | 31% | 417 | 1,1 | 40,1 | 1503 | 31% | 417 | 1,1 | 40,1 | 1503 | 99% | 3 | 0,4 | 6,5 | 1619 |
| **LIAN-5**  | 93% | 41 | 0,6 | 8,7 | 1634 | 98% | 12 | 0,5 | 6,3 | 1617 | 98% | 11 | 0,5 | 6,1 | 1611 |
| **LIAN-10** | 86% | 86 | 1,1 | 10,3 | 1632 | 90% | 65 | 1,1 | 7,8 | 1619 | 92% | 52 | 0,9 | 6,3 | 1610 |
| **LIAN-20** | 66% | 209 | 2,7 | 12,7 | 1627 | 72% | 171 | 1,4 | 7,8 | 1625 | 79% | 130 | 1,6 | 7,8 | 1628 |
| **Theta*-LA** | 4% | 581 | 0,8 | 16,1 | 1454 | 12% | 536 | 2,1 | 37,5 | 1574 | 31% | 421 | 2,2 | 47,4 | 1580 |
| **wTheta*-LA** | 14% | 522 | 2,1 | 35,9 | 1504 | 55% | 277 | 2,76 | 58,3 | 1598 | 73% | 165 | 2,7 | 61,0 | 1567 |

*$m$ is expressed in kilonodes, 1 kilonode = 1 000 nodes

**Fig. 5.** LIAN, Theta*-LA and wTheta*-LA results.

As one can see Theta*-LA is totally inapplicable to angle-constrained path planning when angle constraint is set to 20°-30°. In that case it fails to solve two thirds (or more) of tasks. Weighting a grid, e.g. using wTheta*-LA, significantly (up to several times) improves the performance. But still, wTheta*-LA successfully handles only 14%-55%-73% of the tasks (for angle constraints 20°, 25°, 30° respectively), while the worst LIAN result, e.g. the result of LIAN-20 is 66%-72%-79% respectively. So, one can say, that in general even the "worst" LIAN is 1,5 times better (in terms of the number of successfully handled tasks) than "the best" wTheta*-LA.

Worth mentioning are the results of LIAN-3. While it solves 99% of tasks when angle limit is 30°, in case the latter is 20°-25° only one third of tasks is solved. It indirectly confirms the hypothesis (see section 3.3) that lower values of Δ should be avoided in general. Setting Δ too high – 20 in our case – also degrades the algorithm performance.

If we now take a closer look at the results of best LIAN instances, e.g LIAN-5 and LIAN-10, and compare them to the best results achieved by limited angle Theta*, e.g. to wTheta*-LA results, and use normalization, we'll get the following picture – see figure 6.

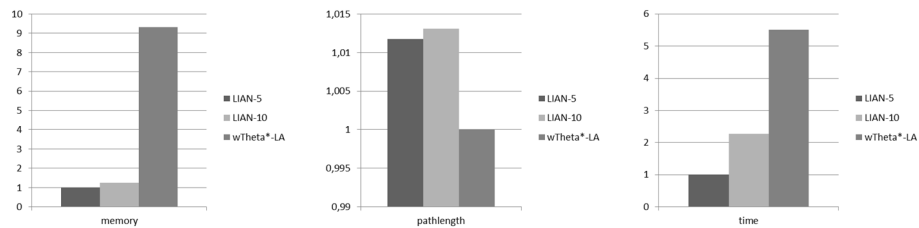

**Fig. 6.** Normalized LIAN-5, LIAN-10, wTheta*-LA results.

As one can see LIAN-5 and LIAN-10 both significantly (up to 5-10 times) outperform wTheta*-LA in terms of time and memory usage. At the same time, path produces by them are only 1% longer that wTheta*-LA paths.

When the best algorithms of LIAN family (e.g. LIAN-5 and LIAN-10) were identified we conducted another experiment, tailored to answer the following question – can their performance be further improved by using dynamic Δ adjustment technique as described in section 3.3? So we repeated 400 tests but now only LIAN-5, LIAN-10 and their dynamic modifications D-LIAN-5, D-LIAN-10 were used (the latter were parameterized as it was suggested in section 3.3). The results are shown in figure 7.

|  | $α_m$ = 20 | | | | | $α_m$ = 25 | | | | | $α_m$ = 30 | | | | |
| --- | --- | --- | --- | --- | --- | --- | --- | --- | --- | --- | --- | --- | --- | --- | --- |
|  | sr | PAR-10 | t | m* | pl | sr | PAR-10 | t | m* | pl | sr | PAR-10 | t | m* | pl |
| **DLIAN-5** | 95% | 34 | 0,8 | 8,9 | 1632 | 98% | 12 | 0,5 | 6,1 | 1616 | 99% | 8 | 0,4 | 6,7 | 1614 |
| **LIAN-5** | 93% | 41 | 0,6 | 8,7 | 1634 | 98% | 12 | 0,5 | 6,3 | 1617 | 98% | 11 | 0,5 | 6,1 | 1611 |
| **DLIAN-10** | 86% | 86 | 1,0 | 8,9 | 1628 | 90% | 59 | 0,8 | 6 | 1624 | 93% | 43 | 1,0 | 5,9 | 1615 |
| **LIAN-10** | 86% | 86 | 1,1 | 10,3 | 1632 | 89% | 65 | 1,1 | 7,8 | 1619 | 92% | 52 | 0,9 | 6,3 | 1610 |

*$m$ is expressed in kilonodes, 1 kilonode = 1 000 nodes

**Fig. 7.** LIAN and D-LIAN results.

As one can see dynamic adjustment of Δ increases the chances of finding a solution. It also decreases running time and memory usage in some cases (for example, when Δ=10, dynamic adjustment reduces the memory consumption on notable 10-15%). So D-LIAN proves to be a worthwhile modification of LIAN.

Summing up all the results one can claim that LIAN (especially with dynamic Δ adjustment, and initial Δ values set to 5 or 10) is an effective algorithm to solve angle constrained path planning problems on square grids and it significantly outperforms it's direct competitors, e.g. wTheta*-LA, in terms of computational efficiency and the ability to accomplish path finding tasks (at least when the urban outdoor navigation scenarios are under consideration).

## 5   Conclusions and future work

We have investigated the angle constrained path planning problem for square grids and presented a new parameterized algorithm – LIAN (and it's variation D-LIAN) – for solving it. We have proved that LIAN is sound and complete (with the respect to it's input parameter – Δ). We have studied LIAN experimentally in various modeled outdoor navigation scenarios and showed that it significantly outperforms existing analogues: it solves more angle constrained path planning tasks than the competitors while using less memory and processing time.

In future we intend to develop more advanced techniques of dynamic Δ adjustment, aimed at further improvement of LIAN performance. Another appealing direction of research is evaluating LIAN in real environments, e.g. implementing LIAN as part of the intelligent control system that automates navigation of a mobile robot or unmanned aerial vehicle in real world.

**Acknowledgements.** This work was partially supported by RFBR, research project No. 15-07-07483.

**References**
1. Lozano-Pérez, T., & Wesley, M. A. 1979. An algorithm for planning collision-free paths among polyhedral obstacles. Communications of the ACM, 22(10), 560-570.
2. Bhattacharya, P., & Gavrilova, M. L. 2008. Roadmap-based path planning-Using the Voronoi diagram for a clearance-based shortest path. Robotics & Automation Magazine, IEEE, 15(2), 58-66.
3. Kallmann, M. 2010. Navigation queries from triangular meshes. In Motion in Games (pp. 230-241)
4. Yap, P. 2002. Grid-based path-finding. In Proceedings of 15th Conference of the Canadian Society for Computational Studies of Intelligence, 44-55. Springer Berlin Heidelberg.
5. Sturtevant, N. R. 2012. Benchmarks for grid-based pathfinding. Computational Intelligence and AI in Games, IEEE Transactions on, 4(2), 144-148.
6. Elfes, A. 1989. Using occupancy grids for mobile robot perception and navigation. Computer, 22(6), 46-57.
7. Dijkstra, E. W. 1959. A note on two problems in connexion with graphs. Numerische mathematik, 1(1), 269-271.
8. Hart, P. E., Nilsson, N. J., & Raphael, B. 1968. A formal basis for the heuristic determination of minimum cost paths. IEEE Transactions on Systems Science and Cybernetics, 4(2), 100-107.
9. Likhachev M., Gordon G., & Thrun S. 2004. ARA*: Anytime A* with Provable Bounds on Sub-Optimality, Advances in Neural Information Processing Systems 16 (NIPS). Cambridge, MA: MIT Press.
10. Botea, A., Muller, M., & Schaeffer, J. 2004. Near optimal hierarchical path finding. Journal of game development, 1(1), 7-28.
11. Likhachev, M., & Stentz, A. 2008, R* Search. In Proceedings of the Twenty-Third AAAI Conference on Artificial Intelligence. Menlo Park, Calif.: AAAI press.
12. Nash, A., Daniel, K., Koenig, S., & Felner, A. 2007. Theta*: Any-Angle Path Planning on Grids. In Proceedings of the National Conference on Artificial Intelligence (Vol. 22, No. 2, p. 1177). Menlo Park, Calif.: AAAI Press.
13. Harabor, D., and Grastien, A. 2011. Online graph pruning for pathfinding on grid maps. In AAAI-11.
14. Kuwata, Y., Karaman, S., Teo, J., Frazzoli, E., How, J. P., & Fiore, G. 2009. Real-time motion planning with applications to autonomous urban driving. Control Systems Technology, IEEE Transactions on, 17(5), 1105-1118.
15. Munoz, P., & Rodriguez-Moreno, M. 2012. Improving efficiency in any-angle path-planning algorithms. In Intelligent Systems (IS), 2012 6th IEEE International Conference (pp. 213-218). IEEE.
16. Kim, H., Kim, D., Shin, J. U., Kim, H., & Myung, H. 2014. Angular rate-constrained path planning algorithm for unmanned surface vehicles. Ocean Engineering, 84, 37-44.
17. Bresenham, J. E. 1965. Algorithm for computer control of a digital plotter. IBM Systems journal, 4(1), 25-30.
18. Pitteway, M. L. V. 1985. Algorithms of conic generation. In Fundamental algorithms for computer graphics, 219-237. Springer Berlin Heidelberg.
19. http://wiki.openstreetmap.org/wiki/Database
20. Hutter, F., Hoos, H. H., Leyton-Brown, K., & Stützle, T. 2009. ParamILS: an automatic algorithm configuration framework. Journal of Artificial Intelligence Research, 36(1), 267-306.